# Benchmarking bias: Expanding clinical AI model card to incorporate bias reporting of social and non-social factors

Carolina A. M. Heming, MD[1], Mohamed Abdalla, PhD[2,10], Shahram Mohanna, PhD[9], Monish Ahluwalia, MSc, MD[3], Linglin Zhang, MS[4], Hari Trivedi, MD[5], MinJae Woo, PhD[4], Benjamin Fine MD, MSc[2,6] Judy Wawira Gichoya, MD[5], Leo Anthony Celi, MD, MPH, MSc[7,8], Laleh Seyyed-Kalantari, PhD[6,9]

1. University of Iowa, Iowa City, IA, US
2. Institute for Better Health, Trillium Health Partners, 100 Queensway West, 6[th] floor, Mississauga, ON L5B 1B8, Canada
3. Kingston Health Sciences Centre, Queen's University, 76 Stuart St., Kingston, ON K7L 2V7, Canada
4. School of Data Science and Analytics, Kennesaw State University, 3391 Town Point Dr NW, Kennesaw, GA 30144, USA
5. Department of Radiology and Imaging Sciences, Emory University, 1364 E Clifton Rd NE, Atlanta, GA 30322, USA
6. Vector Institute of Artificial Intelligence, 661 University Ave Suite 710, Toronto, ON M5G 1M1, Canada
7. Massachusetts Institute of Technology, 77 Massachusetts Avenue Cambridge, MA 02139, USA
8. Harvard Medical School, 25 Shattuck St, Boston, MA 02115, USA
9. Department of Electrical Engineering and Computer Science, York University, 4700 Keele St, Toronto, Ontario, ON M3J 1P3, Canada
10. Department of Medicine, University of Alberta, Walter C Mackenzie Health Sciences Centre, 8440 112 St. NW Edmonton, Alberta

**Abstract:**

The model facts labels are often used to communicate artificial intelligence (AI) model performance to the end users for proper deployment and often report the overall performance averaged across all populations. However, the overall performance is not generalizable to different subgroups of patients. In this paper, we highlight the importance of currently unforeseen aspects, essential for AI models in healthcare, that are not included in the model fact cards. We suggest including through bias and fairness analysis outcomes in model fact labels in the health care AI model including: (a) socio-demographic fairness factors (e.g., sexes, race, socioeconomic status, etc.) and (b) Technical, Environmental and Biological Bias Factors (e.g. disease-dependent, anatomical, or instrumental factors, etc.). The socio-demographic fairness factors have been explored in fairness analysis of AI models in healthcare but have not often been listed in model fact labels. Besides, the technical, environmental, and biological bias factors have not often been explored in the bias analysis of AI models in healthcare and have not been reported in model fact cards. Here, we demonstrate disparities with respect to both sets of factors. We suggest that clinical AI model reporting cards need to be expanded to communicate such a wider range of disparities to end users for precise patient care. Analysis and communication of such a wider range of biases identify potential drivers of bias, which pave the way to debiasing. As well, it highlights the model caveats and guides the end user and policymaker on how, when, and under what circumstances the model works or fails.

## 1- Introduction:

Artificial intelligence (AI) models in healthcare often achieve performance on par with human specialists [1,2]. However, AI models with high performance on the overall population may have performance disparities for specific sub-populations, which has been noted as a bias problem in such models.

Discrepancies in AI model performance between sub-populations have been widely demonstrated in many applications. For example, race and sex bias has been reported in AI models developed for medical image disease diagnosis [3]. Patients' self-reported race can be detected from their medical images alone [4] by AI algorithms, and it is known that some algorithms may also have worse performance for historically underserved races (e.g. Hispanic or Black patients). Such biases in healthcare AI-based decision-making tools are a critical issue that must be detected and, if not mitigated, at least communicated to the end user and addressed before large-scale adoption.

The model facts labels are proposed to report clinical AI model performance to end users e.g., clinical staff [5]. However, these model cards often rely on overall performance reports, which do not demonstrate the full picture of how, when, and under what circumstances the model works or fails. Mitchell et al. introduce a model card for non-healthcare applications, which includes fairness analysis [6] and demonstrates the impact of instrumental on environmental factors in model bias and fairness. Due to the potential critical direct impact of healthcare AI model failures on human lives, thorough bias and fairness analysis in the model card is crucial.

A "Model Card" or "Model Facts Label", is a systematic effort to ensure that front-line clinicians know how, when, how not, and when not to incorporate model output into clinical decisions [5]. The major sections of a "Model Facts Label" include the model's name, location, and version, summary of the model, mechanism of risk score calculation, validation and performance, uses and directions, warnings, and other information [5, 6, 31, 32, 43]. Table 1 shows the specification of AI model cards in healthcare, including commonly addressed items and unforeseen items which need to be considered.

*Table 1: Specification of AI model cards in healthcare, including items and unforeseen criteria*

|  | **Items mostly used [5, 6, 31, 32, 43]** | **Unforeseen items [10-12, 36, 40-42]** |
|---|---|---|
| **Existing model fact cards in healthcare** | -Model's details, name, location, version, date, use the mechanism of risk score calculation, validation and performance, uses and directions, warnings and licenses.<br><br>– Information about intended use, training datasets, preprocessing, evaluation factors, training algorithms, validation approaches and parameters, fairness constraints or features, ethical considerations, caveats and recommendations. | No subgroup analysis of performance, and only report outcomes on the overall population<br><br>Fairness analysis on socio-demographic factors (e.g. sex, race, socioeconomic status, etc.)<br>Bias analysis:<br>1- Technical factors<br>   e.g., imaging devices<br>2- Environmental factors:<br>   Source of data (e.g. department)<br>3- Biological factors: |

|  |  | e.g anatomic factors, disease dependence factors |  |
|--|--|--|--|
|  |  |  |  |

Implied in Table 1, the importance of bias and fairness analysis and communicating the outcomes in model fact labels in the AI model has been overlooked. This is critical due to the lack of fairness widely shown in the literature [10-12, 36, 40-42].

A broader investigation is required beyond known socio-demographic fairness factors in bias analysis, and desire reflection in model cards, which we call technical, environmental, and biological bias factors. This category highlights the disparate outcome of AI models across (1) anatomic factors (e.g., body habitus, anatomic variants), (2) disease-dependent factors (e.g., disease appearance), (3) instrumental factors (e.g., imaging devices), and (4) data sources. This topic has not been explored much in the healthcare domain and has not been reported in model cards, which we have introduced and discussed in more detail in this paper. In the end, we propose a comprehensive AI Model fact label for medical imaging that incorporates bias and fairness investigation. This is important to communicate with end users, policymakers and for AI regulation.

*Reviewing the topics for Model Card Documentation*

A scientific review was done of publications during 2018-2024, with the keywords: "Model Card", "AI Model Card", "AI model fact sheet", "AI model fact sheet in medical imaging" and "AI fairness in model fact sheet", focusing on bias and fairness applicable to healthcare and medical imaging. Some works are reported as standardization of AI model cards in ophthalmology for improving transparency [26], enhancing reproducibility of machine learning-based studies through model cards [27], evaluating the effectiveness and perception of information quality of the Model Cards [28], reducing racial bias in AI models for clinical use requires a top-down intervention [29], the importance of internal and external validation for evaluating machine learning (ML) models before deployment [30], measuring model biases in the absence of ground truth [31] and characterization of AI model configurations for model reuse [32].

In a non-medical application [3, 6], the authors suggested adding fairness analysis per subgroup of sexes, ages, or races (socio-demographic fairness factors) in the model fact sheet as the performance is inconsistent across groups considering these factors. However, model fact labels in healthcare do not often include fairness analysis outcomes and normally report the overall performance [5, 31, 32], while the overall performance report does not demonstrate the full picture of how, when, and under what circumstances the model may work or fail [5, 31, 32].

There is no comprehensive bias consideration i.e. in [31] only "gender biased" labels have been used. Exploring the literature, none of the existing proposed model cards for healthcare applications have discussed the importance of reporting through fairness analysis with respect to socio-demographic fairness factors or explored the importance of technical, environmental, and biological bias factors specific to medical imaging. In this paper, we demonstrate disparities with respect to both sets of factors.

*Essential Factors in fairness and bias analysis*

Recognizing and addressing various sources of bias is essential for algorithmic de-biasing and trustworthiness and to contribute to a just and equitable deployment of AI in medical imaging [35, 36]. Biases in clinical deployment of AI may impede their intended function and potentially exacerbate inequities. Specifically, medical imaging AI can propagate or amplify biases introduced in the many steps from model inception to deployment, resulting in a systematic difference in the treatment of different groups [35, 36]. Inherent bias in an AI-model could bring inaccuracies and variabilities during clinical deployment of the model [37]. It is challenging to recognize the source of bias in AI-model due to variations in datasets and black-box nature of system design and there is no distinct process to identify the potential source of bias in the AI-model [37]. It has been indicated that most of the studies in healthcare suffer from data bias and algorithmic bias due to incomplete specifications mentioned in the design protocol and weak AI design exploited for prediction [37].

The bias and fairness analysis paper often explores socio-demographic fairness factors where the main focus of literature in bias and fairness analysis has been on races and sexes [4, 7, 8, 12, 44-46], with less attention to other important factors that cause disparity, such as spoken language [14], education [13], income level [13], age [3, 11], or insurance type [3, 11].

Such a narrow focus misses other sources of bias and does not consider heterogeneity within members of a specific race/sex subgroup i.e. the case of a commercial algorithm that identified black patients as having fewer healthcare needs than white patients with similar medical conditions. This occurred because the model used healthcare costs as a proxy to health status and black patients often spent less on healthcare than white patients due to socioeconomic reasons [8]. If this model was debiased simply by correcting predictions for black patients, the model would still harm low-income patients of all other races since income level cannot be universally attributed to race. Therefore, race is the incorrect bias factor to correct for this instance.

### 2- A comprehensive AI Model fact label for medical imaging - incorporating bias

In this paper, we have suggested a more comprehensive set of factors to be reported in model fact cards in the context of medical imaging. We proposed incorporating a thorough bias and fairness analysis of AI in medical imaging considering disparate outcomes across:

(a) socio-demographic fairness factors (e.g., sex, race, socioeconomic status) in a broader range vs conventional focus on sex and race.

(b) technical, environmental, and biological bias factors (e.g., disease-dependent, anatomical, instrumental factors).

While socio-demographic fairness factors are well introduced, the technical, environmental, and biological bias factors may include (1) **anatomic factors** (e.g., body habitus, anatomic variants), (2) **disease-dependent factors** (e.g., disease appearance), (3) **instrumental factors** (e.g., imaging devices), and (4) **data sources**. Expanding bias reporting to these factors reveals other hidden disparity drivers and allows clear and accurate communication of model biases and limitations to the end user and policymakers.

Indicated in Table 1, socio-demographic fairness factors have been studied in the fairness analysis of healthcare AI models; they are often excluded from model fact labels. Moreover, technical, environmental, and biological bias factors are rarely examined in bias analysis and are not reported in model fact cards.

In the next section, we expand the proposed idea in more detail and apply it to two different use cases in medical imaging.

### *The impact of socio-demographic fairness factors on disparate AI outcomes*

Equality in model performance across subpopulations of a given sensitive attribute has been the focus of bias analysis [3, 11, 12]. The impact of socio-demographic fairness factors such as patient race and sex on disparate outcomes of AI models in health care has been widely demonstrated [4, 7, 8, 13]. However, much less attention to disparate outcomes of AI models with respect to other social factors such as patients' language (English vs. non-English speaker) [14], education [13], income level [13], age [3], or insurance type [3] have been demonstrated. For instance, in a systematic study of AI-based chest X-ray prediction, Seyyed-Kalantari *et al.* [3] found that AI models underdiagnosed historically under-served patients, e.g., such as younger patients or patients with medical insurance type who are often low-income at a higher rate. Relatedly, Zhang *et al.* perpetuated undesired biases, resulting in performance discrepancies with respect to patients' spoken language (English vs. non-English), ethnicity, and insurance type [14]. Besides, Pierson *et al* demonstrate bias with respect to the patients' education and income level in pain and disease severity measures [13]. Such studies demonstrate the importance of widely expanding the socially sensitive attributes to include such factors as described. Fig. 1 and Fig. 2, demonstrate the disparate outcome of both AI models for chest X-ray and mammography abnormality classification task (normal/ abnormal) across sex, age and race subject to the data availability. We denote to report these quantities for all other proposed factors (e.g. Language, socioeconomic status, etc.), but we have been limited by the data availability for some factors.

### *Technical, environmental and biological bias factors impact disparate AI model outcome*

Technical, environmental, and biological bias factors can contribute to AI model bias and are often ignored and has not been much explored, which results in disparate outcomes and bias for groups of patients. These factors may have a great impact on end-users' ability to account for the predictions of AI models in their specific practice. For example, if an AI model constantly has lower performance on patients whose image is gathered using a specific imaging device [9], then the end-user can be informed about this shortcoming in the model card and rely on the outcome of the AI model accordingly. Here, we list these factors that we suggest should be considered in the context of AI models for radiology applications.

1. <u>Anatomic factors:</u>

Anatomic variants and prior conditions may cause errors in diagnosis and treatment planning. For example, anatomic variants in the spine may be a source of inaccurate radiotherapy planning [15]. Similarly, anatomic variants can also be a source of inaccuracy in machine-learning models for organ or tumor segmentation [16]. Oakden-Rayner *et al.* found that a classifier for hip fracture detection on frontal X-rays performed worst in cases with abnormal bone or joint appearance, such as Paget´s disease of the bone [17]. Additionally, differences in body habitus may affect model predictions. For instance, increased breast density is an independent risk factor for breast cancer [18] and simultaneously reduces mammography´s sensitivity in breast cancer screening [19].

As an illustrative case study, a disparate AI model performance across breast density is shown in Fig. 2 [10]. As can be seen, the disparate outcome for BI-RADS density A, B, C, and D where lower AUC for patients with BI-RADS density C, D is visible. In practice, Asian and Black patients have a higher breast density [20] which may result in disparities for AI models due to their breast density, not race. In this case, adjusting model predictions based on race to reduce bias would be inaccurate for two reasons: (1) Asian or Black women with low breast density are now at higher risk of false positives, and (2) White women with high breast density will continue to be underdiagnosed. In this case, rather than adjusting for race, predictions should be adjusted based on breast density.

2. <u>Disease-dependent factors:</u>

The same disease, but with different expressions, may also affect bias in AI model performance. COVID-19 pneumonia may have different expressions depending on the viral variant, immunization status, and phase of the disease (early infection, pulmonary phase, or hyperinflammatory phase) [21]. For instance, studies show that in hospitalized patients with COVID-19, CT was more likely to be negative for pneumonia during periods of Omicron versus Delta variant prevalence and that the proportion of patients with an atypical CT pattern was higher in the Omicron variant group than in the Delta variant group [21]. Therefore, models developed to diagnose COVID-19 pneumonia should consider those variables that not underdiagnosed people with less severe disease. For example, in screening mammograms for a model trained to distinguish between abnormal/normal tissue patches and randomly selected normal tissue patches, the AI model has disparate outcome cross-images with different findings, including mass, architectural distortion, calcification and asymmetry [10]. The performance is lower for those with architectural distortion findings.

Another factor that influences different expressions for the same disease is patient immunity, whether immunocompetent or immunocompromised. It is known that pulmonary tuberculosis also may present with different chest CT findings in HIV patients in comparison to immunocompetent patients [22]. Even different types of immunodeficiency may lead to variability in disease expression. For instance, the pulmonary manifestations of *Pneumocystis jirovecii* pneumonia might have different chest CT patterns among HIV-positive patients compared to patients with other reasons for immunosuppression, as in hemato-oncologic and post-transplant patients and in patients under immunosuppressive drugs due to autoimmune diseases [23]. Those differences, if not recognized, may be a source of disparity in the correct diagnosis of patients that require more complex care.

Variable tumor appearance may also impact model performance in specific subgroups. For example, pancreatic adenocarcinoma may be iso-attenuating in up to 5.4% of cases making them visually indistinguishable from the surrounding pancreatic parenchyma on dynamic CT imaging and difficult to diagnose [24]. Therefore, the consistently lower sensitivity of an AI model for specific tumor appearance could result in disparate outcomes. In the case study (sensitivity in detecting each pathology class) shown in Fig. 1, the performance of the AI model in normal/abnormal classification of chest radiographs has been plotted when there is a solitary finding vs when there are more than one (two or more pathology classes) abnormal category in chest radiographs [9]. It is visible that the classifier performance was universally lower in one clinically important scenario, i.e. cases with a solitary finding. with two or more pathology classes.

3. Instrumental factors:

AI model performance for face detection may vary depending on what cameras are used [6]. This is also the case in medical imaging: Ahluwalia et al. observed that the performance of AI models trained for abnormality classification varied substantially across different imaging devices in radiology [9. 19]. For example, as shown in Fig. 1, there is a 23% difference in sensitivity and specificity when applied to images taken using the GE Type 1 compared to images taken using a Varian Type 1. Similar disparate outcomes for AI models trained for mammography screening have been observed across imaging devices (See model card in Figure 2, ΔAUC and ΔF1 score across imaging devices). Documenting these details in a "model fact card" guides the end-user on the potential disparate performance based on the imaging device that they use in practice. As a result, it is important to include this factor in a model fact card.

4. Data source:

Data sources (i.e., the type of hospital and its patient demographic and medical conditions) impact model performance. Additionally, model performance differs even across multiple departments within the same hospital. These categories may also be called environmental factors as it depends on where the image is gathered and under what environment it has been taken. For instance, as previously shown models trained on the CheXpert [25] dataset, which has more tertiary care center cases, have less bias [3] than models trained on Chest-Xray, which is gathered from a hospital that does not do routine procedures [3]. External validation of four AI models trained on four different datasets for the same disease classification task demonstrated reduced sensitivity, but increased specificity, in emergency room patients. The reverse was true for inpatients and ICU patients [9] (See model card in Fig. 1 for ΔM across different departments within the same hospital). Similarly, for mammography screening AI models, we can find the disparate outcome of AI models across different departments in the hospital (See Fig. 2). The disparate outcomes across data sources must be evaluated and communicated to the end user to ensure appropriate, safe, and fair implementation of AI models into clinical care.

*Model Validation and Verification*

Clinical use of AI tools requires rigorous validation and verification to ensure their safety, effectiveness, and reliability. However, these processes can be challenging due to the high variability in medical imaging data and clinical scenarios. The choice of 'Metrics' to report and subgroups (i.e. 'Factors') we choose to report the outcome depends on the domain expert. However, we suggest a thorough analysis of a diverse subset of applicable factors to ensure we are not missing hidden existing biases.

Furthermore, models that perform well during validation may still fail in real-world applications due to the domain shift or model drift, which refers to the changing data distributions over time or across different settings. Domain shift can occur when the data used to train an AI model differs from the data used to train it. Also, Model drift can occur over time as the data distribution changes. It is important to use a rigorous validation and verification process for AI tools to address these challenges at every step, including Data collection, Model evaluation and Model monitoring [38]. As a result, in the model fact label, it is important to communicate the training and evaluation data and inform the end user that the reported evaluation metrics are valid within the specified version of the model and training and testing data. The sections of 'Model Details', 'Intended Use', 'Training & Evaluation data' and 'Caveats & Recommendations' in the model card in Fig 1 and Fing 2 are covering these aspects. These sections try to communicate the model task and usage specification, version of the training and test data, etc. Then the quantitative analysis and the graphs of the subgroup analysis highlight the outcome of the reported results, overall outcome and the disparate outcome across subpopulations.

**Experiment**:

Here we use two demonstrative examples to demonstrate disparate outcomes across a wide range of factors that we mentioned, and we communicate those disparities using model fact labels. We should note that data availability was limited when selecting the factors and not all the mentioned factors are applicable or available for different use cases. Fig. 1 shows the model card of an AI model trained on the CheXpert dataset, externally validated on a dataset of 200,000 chest x-rays from a tertiary care center [9]. Fig. 2 demonstrates the model card for the imaging abnormality classification in screening mammogram analysis [10]. Model cards often include model details and information on what data the model has been trained on. In addition, for a given metric M∈{Accuracy, F1 score, sensitivity, specificity, AUC, …}, we have reported ΔM = $M_{subgroup}$ - $M_{overall}$, which demonstrates how much gap the subgroup measure of metric M, $M_{subgroup}$, is experiencing compared to the overall population, $M_{overall}$. A positive gap means the model performs in favor of the given subgroup, while the negative gap demonstrates the subgroup is unfavorable.

# Model Facts Card – Chest Radiograph Classification

**Model Details**
- State-of-the-art convolutional neural network with a 121-layer DenseNet architecture developed by Kalantari et al., 2020 based on the CheXpert Image dataset
- Classifies chest radiographs into one or more of 14 classes: atelectasis, cardiomegaly, consolidation, edema, enlarged cardiomediastinum, fracture, lung lesion, lung opacity, pleural effusion, pleural other, pneumonia, pneumothorax, support devices, and no finding
- The no finding category was used to determine binary classes (normal if 1, abnormal if 0)

**Intended Use:**
- Chest radiograph classification for research purposes only

**Factors:**
- **Groups** include age, EHR-reported sex, and name-based ancestry (Greater European, Greater African/Indian, and Greater East Asian), and 12 pathology classes
- **Instrumentation factors** include 8 chest radiograph imaging systems
- **Environment factors** include patient location (emergency room, inpatient, outpatient, and ICU)
- Other **relevant factors** not studied include image quality and rotation

**Metrics:**
- **Evaluation metrics** include accuracy, positive predictive value (PPV), negative predictive value (NPV), sensitivity, and specificity for each factor and group; sensitivity is provided for pathology classes
- Metrics were calculated from respective confusion matrices
- Bootstrap resampling (n = 10,000) was used to generate 95% confidence intervals

**Training & Evaluation Data:**
- **Trained** on 224,316 chest radiographs from Stanford University Medical Center between October 2002 and July 2017 (Irvin et al., 2019); no retraining nor threshold adjustment was performed
- **Evaluated** on 197,540 chest radiographs from Trillium Health Partners from January 2016 to December 2020 (Ahluwalia et al., 2023)

**Caveats & Recommendations:**
- Only applicable to Trillium Health Partners

**Quantitative Analysis:**
- Accuracy of 76% in binary classification with skewing to higher specificity vs. sensitivity
- Higher sensitivity in detecting pleural effusions; lower sensitivity in detecting cardiomegaly, fracture, and all pathologies when solitary
- Low sensitivity and high specificity in patients under 40 and in the emergency room
- High sensitivity and low specificity in patients over 65 and in the ICU
- Similar performance relative to sex and ancestry; variable performance on different equipment models

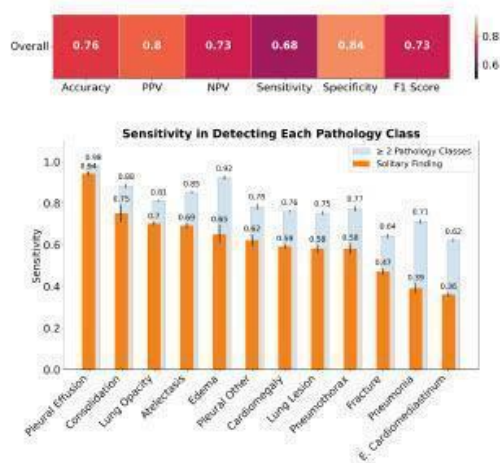

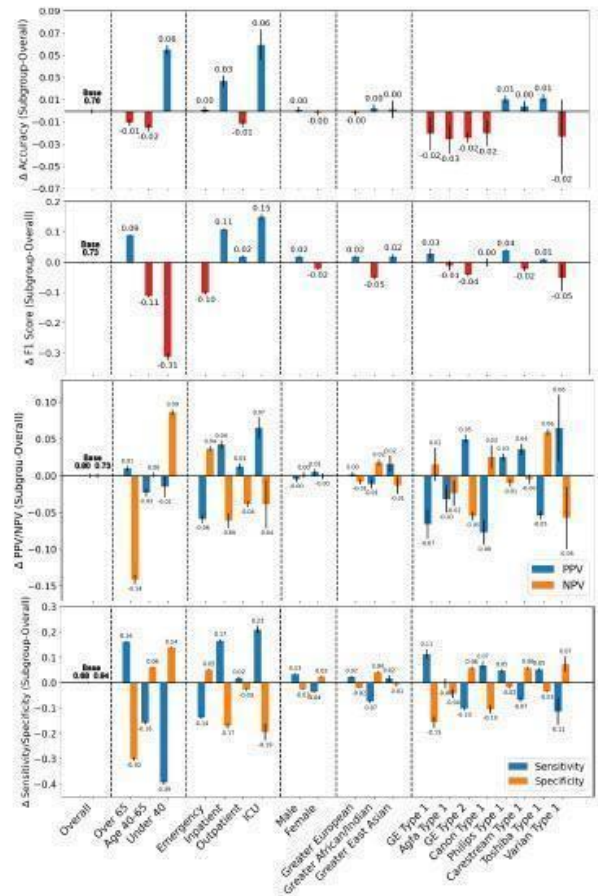

Fig.1 Example of suggested model card showing model performance analysis across social and non-social factors. In this case we analyzed a third-party classifier applied to 200,000 images from a tertiary care center [9]. We plot the difference in performance as measured by multiple metrics to demonstrate disparities.

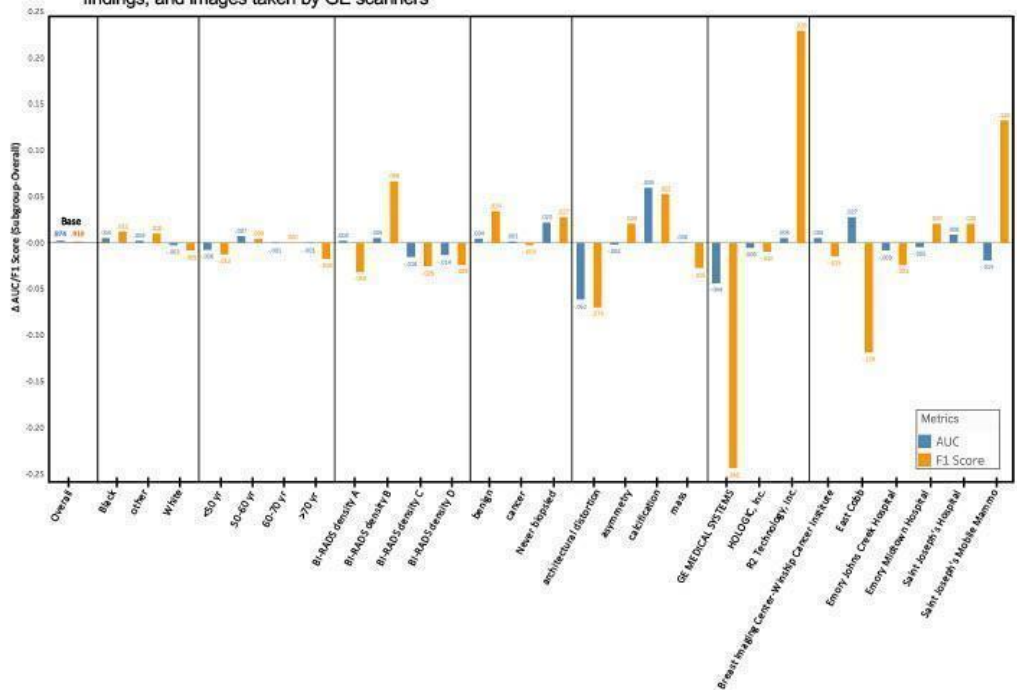

## Model Facts Card – Abnormality Classification of Screening Mammograms

**Model Details**
- A deep convolutional neural network enhanced by ResNet152V2 (He et al., 2016) pretrained on ImageNet (Deng et al., 2009), six additional dense layers were added
- Classifies screening mammogram patches as normal (BIRADS 1 or 2) or abnormal (BIRADS 0)

**Intended Use:**
- Screening mammograms classification for research purpose only

**Factors:**
- **Groups** include race (White, Black, other races), age, and 3 pathology classes (never biopsied, benign, and cancer)
- **Anatomic** tissue density
- **Image Findings** include mass, asymmetry, architectural distortion (AD), and calcification
- **Instrumental Factors** include 3 mammography device manufacturers
- **Environmental Factors** include screening mammography exam location

**Metrics:**
- Evaluation metrics include accuracy, area under the receiver operating characteristic curve (AUC), recall, precision, F1 score
- Bootstrap resampling ($n \in [1000, 13390]$) was used to generate 95% confidence interval (CI)

**Training & Evaluation Data:**
- 52,444 mammogram patches extracted from Emory Breast Imaging Dataset (EMBED) between 2013 to 2020 (Jeong et al., 2023)

**Caveats & Recommendations:**
- Does not represent performance at institutions outside Emory University
- Results may not be generalizable to other racial groups beside White and Black

**Quantitative Analysis:**
- Accuracy of 92.6% (95% CI = 92.0–93.2%), AUC of 0.975 (95% CI = 0.972–0.978) in binary patch classification of abnormality
- Patch classification performance was similar across race, age, and pathologic outcome; decreased performance was observed in dense breasts (density D), patches with architectural distortion (AD) compared to other imaging findings, and images taken by GE scanners

*Fig 2. Model card for classification of abnormalities in screening mammography. Statistical analysis indicates that there are both social and non-social factors, such as anatomic factor (breast density), disease-dependent factors (architectural distortion), and instrumental factors, aggravating classification performance, which may lead to failure of the abnormality of object detection in mammograms [10].*

### 3- Discussion:

In the absence of effective debiasing techniques for AI medical imaging, it is crucial to perform a thorough analysis of model bias across diverse factors and communicate the existing biases to the end users, domain experts and policymakers, this approach is essential because biases in medical imaging AI can significantly impact clinical decision-making and patient outcomes, potentially exacerbating health disparities among different patients. By analyzing biases related to both socio-demographic fairness factors and technical, environmental and biological bias factors, healthcare providers can better understand the limitations and risks associated with the proposed AI model. Also, developers can better diagnose the potential root causes of the bias. Furthermore, transparent communication of these biases to end users ensures that clinicians are aware of the AI system's potential shortcomings and can make more informed decisions, thereby mitigating some of the adverse effects of biased AI models [39-42].

In this paper, we proposed a framework for model fact labels considering the challenges specific to medical imaging. The former model card often includes information related to the training and test data of the AI model as well as information on the general performance of the model across all populations on the test set. The purpose is to inform the end user about the potential expectation of the model behavior on the test cases that it may see in practice. In our framework, we suggest the need for expanding the model performance report such that it includes subgroup analysis. Therefore, the end user will have a better understanding of the strengths and weaknesses of the model and potential points of failure or lower performance. These caveats are detectable through fairness and bias analysis.

We demonstrate that the fairness and bias sensitivity of the AI model are not limited to the known demographic features of the patients. Considering the use cases of medical imaging we discussed many other technical, environmental and biological factors that can impact the model's disparate outcome. As illustrative examples, we show a model card label for two different use cases: firstly, an AI model for X-ray normal/abnormal classification and the second model in mammogram screening normal/abnormal classification. We communicated the biases of these models with respect to both socio-demographic fairness factors and technical, environmental and biological factors to the end user in model fact labels. Our examples also confirmed the disparate outcome specifically with respect to technical, environmental and biological factors, which are often ignored in the literature. Such detailed investigation of model biases and communicating them with the end user is important for the accountability of the model under different technical, environmental, and biological bias factors.

While we have discussed many factors that can bias a model, we should note that all models are not impacted by all these factors. In practice, first, we are limited by available data. For example, if the race data is not gathered initially, we cannot do the bias investigation with respect to race factor. This is essential in the phase of data gathering to include all the relevant features that can possibly be collected (e.g., imaging device) in the original data set. Second, per application, one factor might be relevant or less important to be considered. Domain experts are the best at reaching out to come up with a proper list of potential factors of interest.

It should also be noted that the model fact card is generated based on a specific train and test data. Therefore, the model fact label should introduce the datasets and the population that the model has been trained on. Then all the reported results should be considered in the proposed setup. We cannot make any further suggestions about generalizing to other setups. Some research highlights the importance of internal and external validation for evaluating machine learning (ML) models before deployment [30]. We propose that it is necessary to perform an external validation and report external validation results in the model fact cards. Each model should be fully analyzed to see if it is good and demonstrate some sort of generalizability of the proposed AI model. While it is a good practice to show the generalizability of the model [33, 34], the outcome of external validation should not be overgeneralized as such results only demonstrate model generalizability on the specified external validation datasets.

We are also aware that the fairness assessment is still performance-based, which means it is still relying on binary outcome and comparisons to the ground truth label. This is not the best metric, since the ground truth is biased. Therefore, having a good grade on such a model card does not guarantee that the algorithm is fair. In the end, the ultimate fairness metric is the impact on downstream clinical outcomes. This ensures that all patient groups receive fair and accurate medical care, addressing systemic inequities and improving public trust in AI technologies.

This work is also crucial for policymakers from different perspectives. Firstly, it highlights the need for comprehensive AI model fact labels in healthcare that include thorough fairness and bias analysis. Such a comprehensive model card is important for healthcare applications as it provides more transparency and accountability in AI deployments. Second, by incorporating detailed fairness and bias analyses across socio-demographic and technical, environmental, and biological factors, policymakers can detect a diverse set of root causes for bias in the AI model and make informed decisions to mitigate biases and promote equitable healthcare outcomes. Finally, and most importantly, understanding these disparities allows for the creation of regulations and guidelines that mandate thorough bias reporting and validation processes, ultimately leading to safer and more effective AI models in practice.

**Conclusions**

We proposed an AI Model fact label for medical imaging, incorporating a thorough bias and fairness analysis of AI in medical imaging. Specifically, we highlight disparate outcomes across; socio-demographic fairness factors (e.g., sex, race, socioeconomic status) in a broader range vs conventional focus on sex and race; as well as technical, environmental, and biological bias factors (e.g., disease-dependent, anatomical, instrumental factors). Analysis and communication of such a wider range of biases identify potential drivers of bias which pave the way to debiasing. Developers and end users may select or consider other factors based on their use case and available data. Moreover, they may reconsider gathering features that are not considered regularly and we have shown they may impact the AI model's performance.

**Author contributions:**
All authors contributed to the creation of this paper.

**Acknowledgements:**
We acknowledge the support of the Natural Sciences and Engineering Research Council of Canada (NSERC) Discovery Grant and Connected Mind Canada First Research Excellence Fund (CFREF) grant to L. S. K. AC is funded by the National Institute of Health through NIBIB R01 EB017205. The funder played no role in study design, data collection, analysis and interpretation of data, or the writing of this manuscript.

**Competing Interests:**
All authors declare no financial or non-financial competing interests.

**Data availability**
The datasets used and/or analyzed during the current study are available from the corresponding author on reasonable request.

**Code availability**
The underlying code for this study and training/validation datasets are not publicly available but may be made available to qualified researchers on reasonable request from the corresponding author.